\documentclass{article} 
\usepackage{iclr2023_conference,times}

\usepackage{amsmath,amsfonts,bm}

\def\eqref#1{equation~\ref{#1}}

\def\1{\bm{1}}

\DeclareMathAlphabet{\mathsfit}{\encodingdefault}{\sfdefault}{m}{sl}
\SetMathAlphabet{\mathsfit}{bold}{\encodingdefault}{\sfdefault}{bx}{n}

\usepackage{hyperref}
\usepackage{url}
\usepackage[skip=1.5pt plus0.5pt, indent=0pt]{parskip}

\usepackage{algorithm}
\usepackage{algorithmic}
\usepackage{multicol}
\usepackage{bbm}

\usepackage{graphicx}
\usepackage{subfigure}
\usepackage{adjustbox}

\usepackage{enumitem}
\setitemize{noitemsep,topsep=0pt,parsep=0pt,partopsep=0pt,leftmargin=*}

\newlength\myindent
\title{Learning to Backdoor Federated Learning}

\author{Henger Li$^{1}$, Chen Wu$^{2}$, Sencun Zhu$^{2}$, and Zizhan Zheng$^{1}$ 
\\
$^{1}$ Department of Computer Science, Tulane University, \texttt{\{hli30, zzheng3\}@tulane.edu}.\\
$^{2}$ Department of Computer Science, Penn State University, \texttt{\{cvw5218, sxz16\}@psu.edu}.
}

\iclrfinalcopy 
\begin{document}

\maketitle

\begin{abstract}

In a federated learning (FL) system, malicious participants can easily embed backdoors into the aggregated model while maintaining the model's performance on the main task. To this end, various defenses have been proposed recently, including training stage aggregation-based defenses and post-training mitigation defenses. While these defenses obtain reasonable performance against existing backdoor attacks, which are mainly heuristics based, we show that they are insufficient in the face of more advanced attacks. In particular, we propose a general reinforcement learning-based backdoor attack framework where the attacker first trains a (non-myopic) attack policy using a simulator built upon its local data and common knowledge on the FL system, which is then applied during actual FL training. Our attack framework is both adaptive and flexible and achieves strong attack performance and durability even under state-of-the-art defenses. Code is available at \url{https://github.com/HengerLi/RLBackdoorFL}.
\end{abstract}

\section{Introduction}

A backdoor attack against a deep learning model is one where a backdoor is embedded into the model at the training stage and is triggered at the test stage only for targeted data samples. Such an attack is practical and difficult to defend due to its stealthiness and flexibility. 
The problem becomes more interesting and challenging (for both the attacker and the defender) in a distributed learning system such as federated learning (FL), where devices collaboratively train a model without sharing their local data~\citep{mcmahan2017communication}. In this setting, a malicious insider can only control its local model and has limited influence on the global model, while the server only has access to the models shared by the devices but not their local data. Despite these difficulties, \cite{bagdasaryan2020backdoor} shows that when there is no defense, even a single malicious insider can inject a backdoor into the global model while maintaining its performance on the main task by first training a backdoored model using local data and then poisoning the global model through model replacement to amplify the backdoor effect. The stealth of the poisoned updates can be further improved by distributing triggers across multiple cooperative malicious devices~\citep{xie2019dba} and by considering edge-case backdoors~\citep{wang2020attack}, making it possible to bypass certain training stage defenses such as norm-bounding~\citep{sun2019can} and weak differentially private~\citep{geyer2017differentially} defenses. 

An important limitation of existing backdoor attacks against federated learning is that they often target specific types of defenses or require a relatively large number of malicious devices to be effective. Further, none of them can compromise post-training defenses~\citep{wu2020mitigating,nguyen2021flame,rieger2022deepsight} as we show in our experiments. The main reason is that these attacks are myopic and largely ignore the impact of potential defenses on (long-term) attack performance. 
One exception is Anticipate \citep{wen2022thinking}, which estimates the impact of malicious updates by simulating the system's response to attacks in the next few steps, assuming federated averaging is used by the server. However, due to the simplifications made, it fails under defenses such as Krum and Median. 

The limited success of existing backdoor attacks might give a false sense of 
robustness of FL systems against backdoors. In this work, we show that there exists a strong backdoor attack that can compromise state-of-the-art defenses with a relatively small number of malicious devices. 
The main idea is to formulate the backdoor attack problem as a Markov decision process (MDP) and utilize the local data of malicious devices and common knowledge to build a simulator of the MDP, that is, a ``world model'' of the real FL environment under attacks. The reward function of the MDP takes into consideration both the main task and the backdoor task. The set of malicious devices first trains an attack policy using deep reinforcement learning (RL) in the simulated environment, which is then applied in the real FL training process. Our RL-based attack framework can be applied to both data poisoning and model poisoning. We focus on the latter in this work due to its flexibility. To reduce the action space, we design a Double Whammy Backdoor Attack (DWBA) method. Rather than crafting a backdoored model from scratch, we first adopt a two-task
learning approach to updating the local model in each FL round and then adjust its weights using ideas from~\cite{zhang2022neurotoxin}. RL is used to optimize the hyperparameters of both steps alternatively. 
Despite action space compression and the gap between the simulated and real environments, 
our RL-based attack achieves high accuracy for both the main task and the backdoor task and better backdoor durability under state-of-the-art training stage aggregation-based defenses and post-training mitigation defenses. Our results indicate the importance of developing more adaptive and stronger defenses against backdoor attacks. 

\section{Reinforcement Learning Backdoor Attack Framework}

\subsection{Background} 
\paragraph{Federated Learning.}
A federated learning system consists of a server and $K$ clients where each client has a local dataset. Let $D_k=\{(x_k^i,y_k^i)_{i=1}^{n_k}\}$ denote the local dataset of the $k$-th client and $n_k = |D_k|$ the size of the dataset.
The empirical loss of a model $w$ for the $k$-th client is defined as $F_k(w)=f(w,D_k):=\frac{1}{n_k}\sum_{i=1}^{n_k} \ell(w, (x_k^i,y_k^i))$, where $\ell(\cdot,\cdot)$ is the loss function. Let $U=\{D_1,D_2,\dots,D_K\}$ denote the collection of all 
local datasets. The goal of federated learning is to find a model $w$ that minimizes the average loss across all the devices: $\min_{w} F(w, U):= \frac{1}{K}\sum_{k=1}^Kf(w, D_k)$. 

\noindent In each federated training round $t$ over $T$ total rounds, the server distributes a global model $w^t_g$ to a set of randomly selected clients $\mathcal{S}^t$ 
and let $\kappa=|\mathcal{S}^t|/K$ denotes the subsampling rate.
Each selected client $k$ then performs $E$ local iterations of stochastic gradient descent to update $w^t_g$ to obtain a local model $w^t_k$ with its own data and sends the model update $g^t_k=w^t_k-w^t_g$ back to the server. 
The server then aggregates the set of model updates $\{g^t_k\}_{k\in \mathcal{S}_t}$ from the selected clients 
using a certain aggregation rule $Aggr$. The global model is then updated as $w^{t+1}_g=w^t_g-Aggr(\{g^t_k\}_{k\in \mathcal{S}^t})$, which is then distributed to the selected clients in the $(t+1)$-th round of training. Various $Aggr$ rules have been proposed in the literature, including simple federated averaging as well as robust aggregation rules such as norm-bounding, Krum, and Median. Further, 
the server may perform a post-training defense $h(\cdot)$ such as Neuron Clipping~\citep{wang2022universal} and Pruning~\citep{wu2020mitigating} at round $T$ on the global model to obtain $\widehat{w}^{T}_g = h(w^{T}_g)$.

\paragraph{Threat Model.}
Without loss of generality, we assume that among the $K$ clients, the first $M$ of them are malicious.  The set of malicious devices is assumed to be fully cooperative. 
In each FL round, each malicious device $k$ (if selected by the server) sends a crafted local update $\widetilde{g}_k^t$ to the server. The goal of malicious devices is to inject a backdoor into the global model so that the model misclassifies any test inputs with a chosen pattern (called a backdoor trigger) embedded to a target label chosen by the attackers (``backdoor task'') while maintaining good performance on clean test inputs (``main task'').  In this work, we assume that the backdoor trigger and the target label are pre-chosen and focus on how to best embed the backdoor into the global model (i.e., model poisoning). To this end, each malicious device $k \in [M]$ first builds a crafted dataset $D'_k$ that is derived from its clean dataset $D_k$ by adding the trigger to a subset of $D_k$. Let $\rho_k=|D'_k|/|D_k|$ denote the poison ratio. Let $U' = (D'_1,...,D'_M)$ denote the collection of poisoned datasets. Then the main task is $\min_w F(w,U)$ as defined above, while the backdoor task is $\min_w F(w,U') := \frac{1}{M}\sum^M_{k=1} f(w,D'_k)$. We define the attacker's goal as $\min_w F'(w):= \lambda F(w,U)+(1-\lambda) F(w,U')$ where $\lambda\in [0,1]$ is a hyperparameter for making a tradeoff between the two tasks. A similar approach has been adopted  in~\cite{bagdasaryan2020backdoor} and~\cite{xie2019dba},

\subsection{Reinforcement Learning based Backdoor Attack Framework}\label{sec:RL}



Unlike previous work that solves the attacker's problem for a single FL round, we consider a backdoor attack as a sequential decision-making problem and formulate it as a Markov decision process (MDP), which can be represented as a tuple $(S, A, T, R, \gamma)$ where
\begin{itemize}
    \item $S$ is the state space. The state at round $t$ is defined as $s^t:=(w^{t}_g,\mathcal{A}^t)$, where $w^t_g$ is the global model parameters 
    and $\mathcal{A}^t$ is the set of  malicious devices sampled by the server, both at round $t$.
    \item $A$ is the action space. Let $a^t:=\{\widetilde{g}_k^t\}_{k=1}^M$ denote the joint action of malicious devices at round $t$. 
    Note that a malicious device not sampled at round $t$ does not send any information to the server; hence its action has no effect on the model update. 
    \item $T : S\times A \rightarrow \mathcal{P}(S)$ is the state transition function, which is jointly determined by the number of benign devices and that of malicious devices, their local datasets and training methods, subsampling rate, and the defense mechanism the server applies. 
    \item $R : S \times A \rightarrow \mathbb{R}_{\leq 0}$ is the reward function. Given state $s^t$ and action $a^t$, we define the expected reward at round $t$ as $r^t:= -\mathbb{E}[F'(\widehat{w}^{t+1}_g)] = -\mathbb{E}[F'(h(w^{t}_g-Aggr(\{\widetilde{g}_i^t\} \cup \{g_j^t\}_{i,j\in \mathcal{S}^t, i \in [M], j\notin [M]})))]$ 
    if $\mathcal{A}^t\neq \emptyset$, and $r^t:=0$ otherwise.
    \item $\gamma \in(0,1)$ is the discount factor for future rewards.
\end{itemize}

The attacker's objective is to find a stationary policy $\pi: S\rightarrow A$ that maximizes the expected total reward 
over $T$ FL rounds, which is equivalent to minimize $\sum_{t=0}^{T-1}\gamma^t \mathbb{E}[F'(\widehat{w}^{t+1}_g)]$. 

\paragraph{Simulated Environment.} To derive the optimal backdoor attack policy, the problem then boils down to solving the MDP defined above. One approach is to apply online reinforcement learning, where the set of malicious devices collaboratively update their attack polices during the interaction with the real FL environment. Given the limited amount of attack opportunities and feedback available, however, this approach is unlikely to work in practice. We instead adopt an offline approach in this work. In particular, we consider the commonly assumed white-box setting where the attacker has prior knowledge about the FL environment including the server's defense mechanism, the local training method, the number of devices, and the subsampling rate.  However, they do not have access to the local datasets of benign devices and/or a distribution learned from privacy leakage attack \citep{li2022learning}. Instead, they use their own data to approximate the local data distributions of other devices. With this information, the attacker can  approximate both the transition function and the reward function of the MDP to build a simulator for it. This can be done before the actual FL training starts or with limited amount of online interaction. With the simulator built, the set of malicious devices can adopt a deep reinforcement learning algorithm such as TD3 \citep{fujimoto2018addressing} or PPO \citep{schulman2017proximal} to solve the MDP to train an attack policy, which is then applied during the actual FL training.


\subsection{Double Whammy Backdoor Attack}\label{sec:DWBA}

For the scalability of our attack against large FL models, we propose to only include in the state the model parameters of the last two hidden layers of $w^t_g$ and the number of attackers sampled at round $t$ (instead of the entire set of attackers).
We further restrict all the malicious devices to take the same action.
To further compress the action space, we consider the following two-step approach for crafting gradients and use reinforcement learning to optimize the hyperparameters of both steps. 
 
\noindent{\bf Local Search.} First, consider a local objective function $\tilde{F}'(w) := \lambda F(w,\{D_k\}_{k \in [M]})+(1-\lambda)F(w,\{D'_k\}_{k \in [M]})$, which is an approximation of $F'(w)$ using attacker's local data (including both clean data and poisoned data pooled together from all malicious devices). At each FL round $t$, the attacker first generates 
a new model $\Tilde{w}^t$ by updating the current global model $w^t_g$ via $E^t$ steps of stochastic gradient descent with respect to $F'(w)$ under 
earning rate $\eta^t$, batch size $B^t$, and poison ratio $\rho^t$ (see Algorithm 1 in the Appendix). 
This step is similar to the method in~\cite{bagdasaryan2020backdoor} and~\cite{xie2019dba}. Let $\Tilde{g}^t=w^t_g-\Tilde{w}^t$ denote the model update. 

\noindent{\bf Model Crafting.} For each layer of $\Tilde{g}^t$, the attacker modifies a fraction $\alpha^t$ of parameters that differ the most from the corresponding parameters of $g^t$, the model update calculated by using attacker's clean data only. In particular, we let $\Tilde{g}^t[i]=\Tilde{g}^t[i]-\beta^t(\Tilde{g}^t[i]-g^t[i])$ for 
each selected coordinate $i$ 
where $\beta^t$ is a one-dimension scaling factor. This $\Tilde{g}^t$ is what each sampled malicious device will send to the server at round $t$. This step is similar to  ~\cite{zhang2022neurotoxin} in spirit. To avoid detection, a small amount of random noise can be added to the crafted gradient at each malicious device. 

The concrete approach used in each of the above steps can be replaced by other algorithms. The key novelty of our approach is that instead of using fixed and hand-crafted hyperparameters, i.e., $a^t_1:=(\rho^t, B^t, E^t, \eta^t)$ in the first step and $a_2^t:=(\alpha^t,\beta^t)$ in the second step, as in existing approaches, we use RL to optimize them. Rather than searching for all the parameters together, the two sub-actions are optimized alternatively. The details are given in the Appendix. 

\section{Experiment}
In this section, we compare our RL-based attack with several state-of-the-art backdoor attacks against federated learning, including BFL~\citep{bagdasaryan2020backdoor}, DBA~\citep{xie2019dba}, PGD attack~\citep{wang2020attack}, Neurotoxin~\citep{zhang2022neurotoxin}, and Anticipate~\citep{wen2022thinking}, under common training stage aggregation-based defenses including FedAvg~\citep{mcmahan2017communication}, Krum~\citep{blanchard2017machine}, Median~\citep{yin2018byzantine}, and norm-bounding~\citep{sun2019can}, and/or post-training mitigation defenses including Neuron Clipping~\citep{wang2022universal} and Pruning~\citep{wu2020mitigating}. Due to the space limit, we moved a detailed description of the experiment setup and additional results to the Appendix. 

\begin{figure*}[t]
    \centering
   \begin{tabular}{ccc}
    \adjustbox{valign=m}{{%
          \centering
          \includegraphics[width=.31\textwidth]{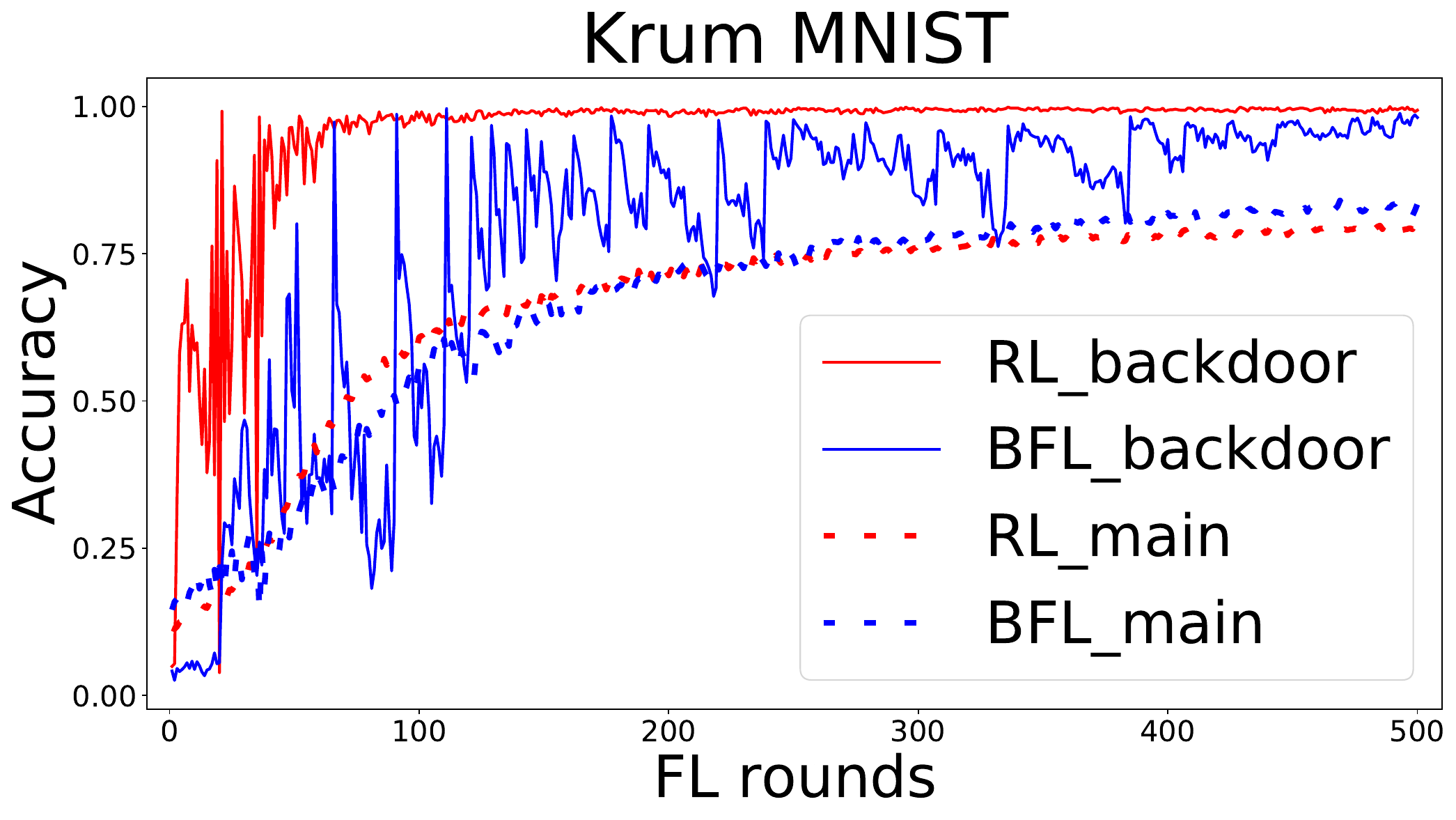}
          }}
         &
    \adjustbox{valign=m}{{%
          \centering
          \includegraphics[width=.31\textwidth]{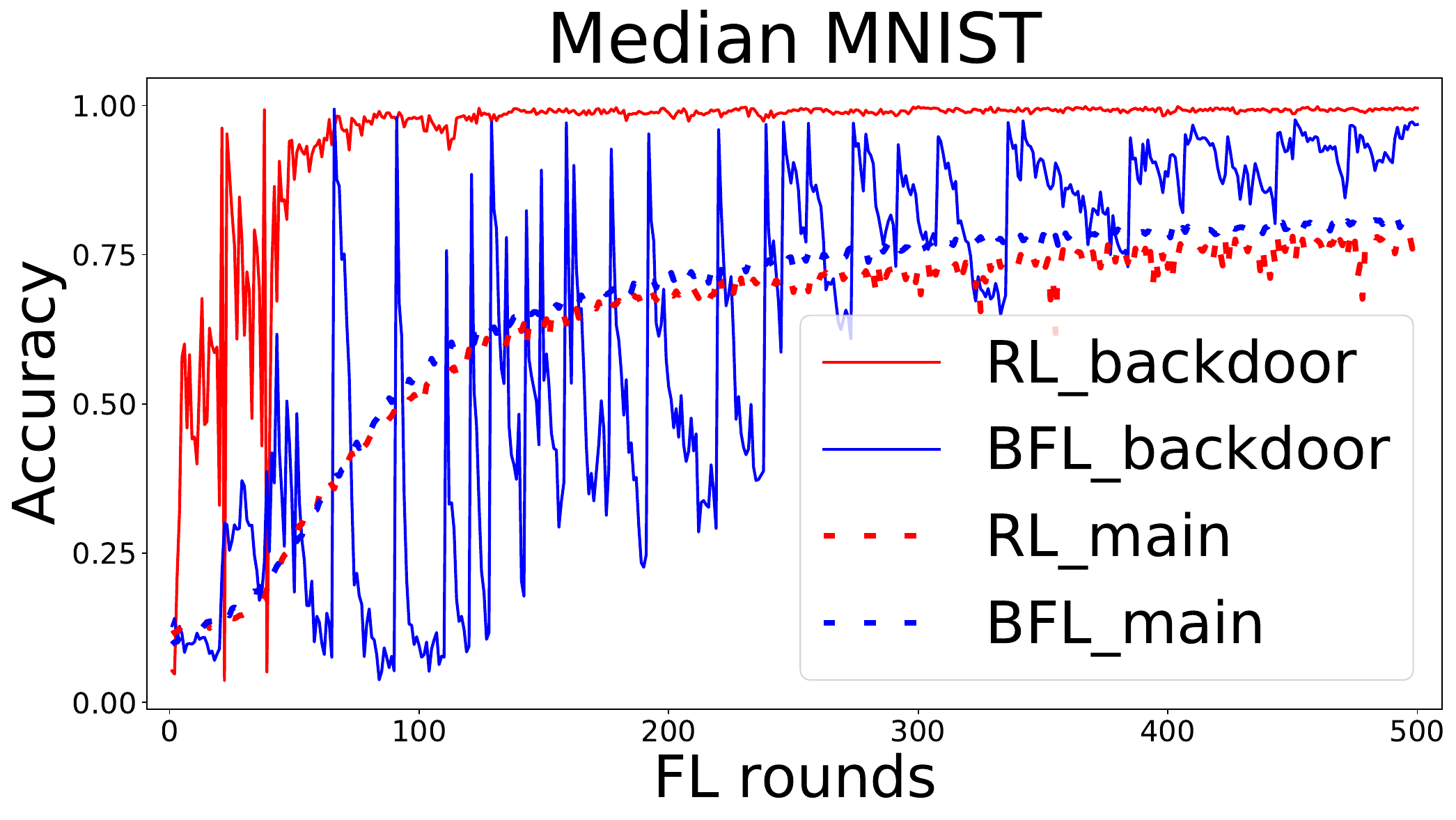}
          }}
         &
    \adjustbox{valign=m}{{%
          \centering
          \includegraphics[width=.31\textwidth]{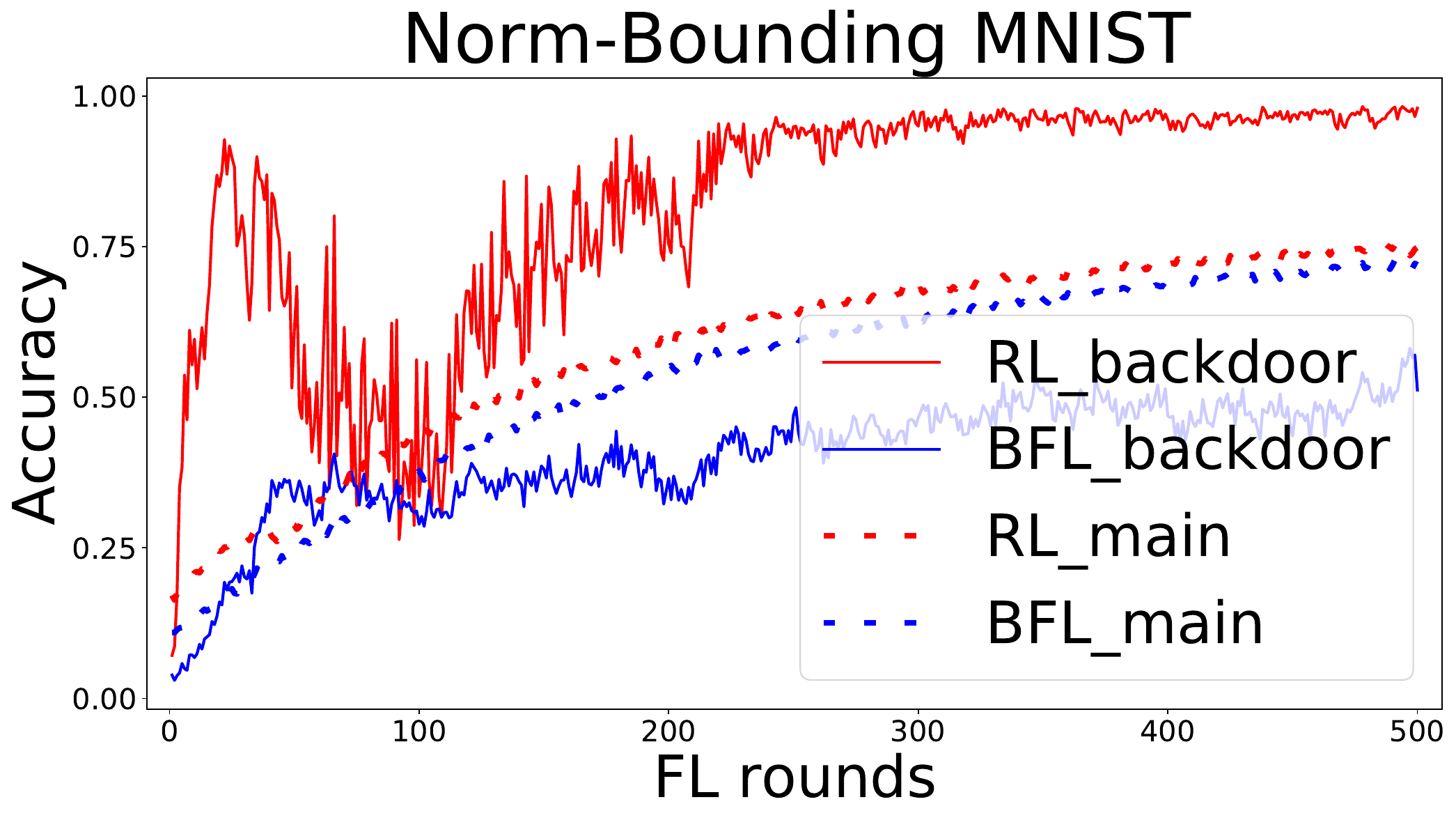}
          }}
          \\
    \end{tabular}
    \caption{\small{Global model accuracy for the backdoor and main tasks during FL training, under BFL and RL attacks, and Krum, Median, and norm-bounding defenses. All parameters are set as default. 
    }}\label{fig:train-stage}
\end{figure*}

\paragraph{Training Stage Defenses.}Figure~\ref{fig:train-stage} shows how the backdoor accuracy and the main-task accuracy of the global model (with respect to the test data) vary over FL training, 
under BFL and our RL attack when the server implements Krum, Median, and norm-bounding defenses, respectively. The default threshold for the norm-bounding defense is 0.05. Results for DBA and PGD under training stage defenses can be found in the Appendix. For both Krum and Median, our RL attack reaches 
$\sim$$100\%$ backdoor accuracy within 100 FL rounds and remains stable afterwards, while the backdoor performance of BFL fluctuates 
during FL training due to subsampling 
and never reaches the same level of high accuracy as the RL attack does. Further, our RL attack maintains similar accuracy for the main task as the baselines. 
For the norm-bounding defense, our RL attack outperforms BFL for both the backdoor and main tasks. In particular, for the backdoor task, the RL attack again reaches $\sim$$100\%$ accuracy after 220 FL rounds, while the backdoor accuracy of BFL slowly grows and never goes beyond 
$50\%$ during FL training. 

\begin{figure*}[t]
    \centering
   \begin{tabular}{ccc}
    \adjustbox{valign=m}{{%
    \subfigure[]{
          \centering
          \includegraphics[width=.31\textwidth]{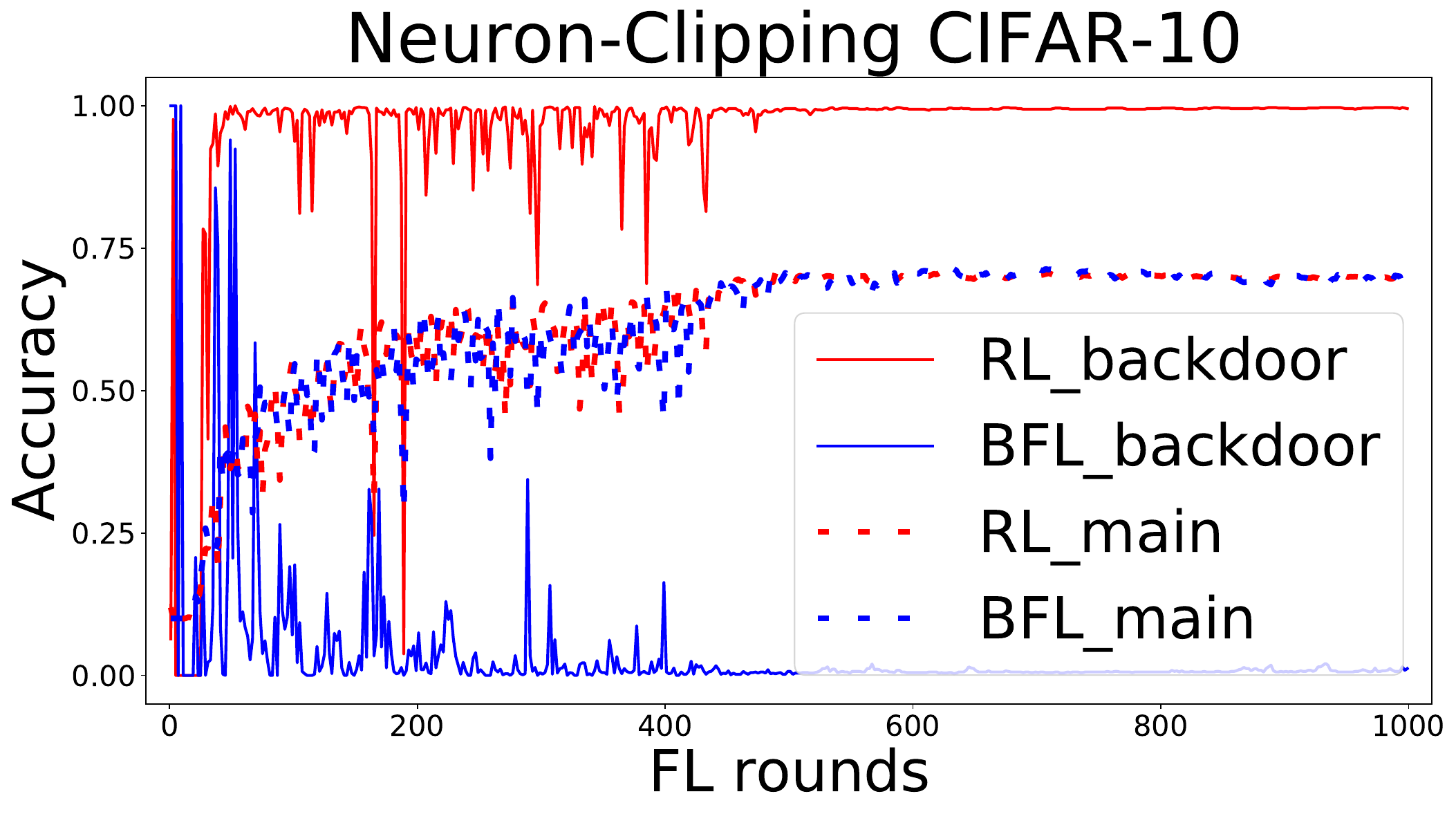}
          }
          }}
         &
    \hspace{-1em}
    \adjustbox{valign=m}{{%
    \subfigure[]{
          \centering
          \includegraphics[width=.31\textwidth]{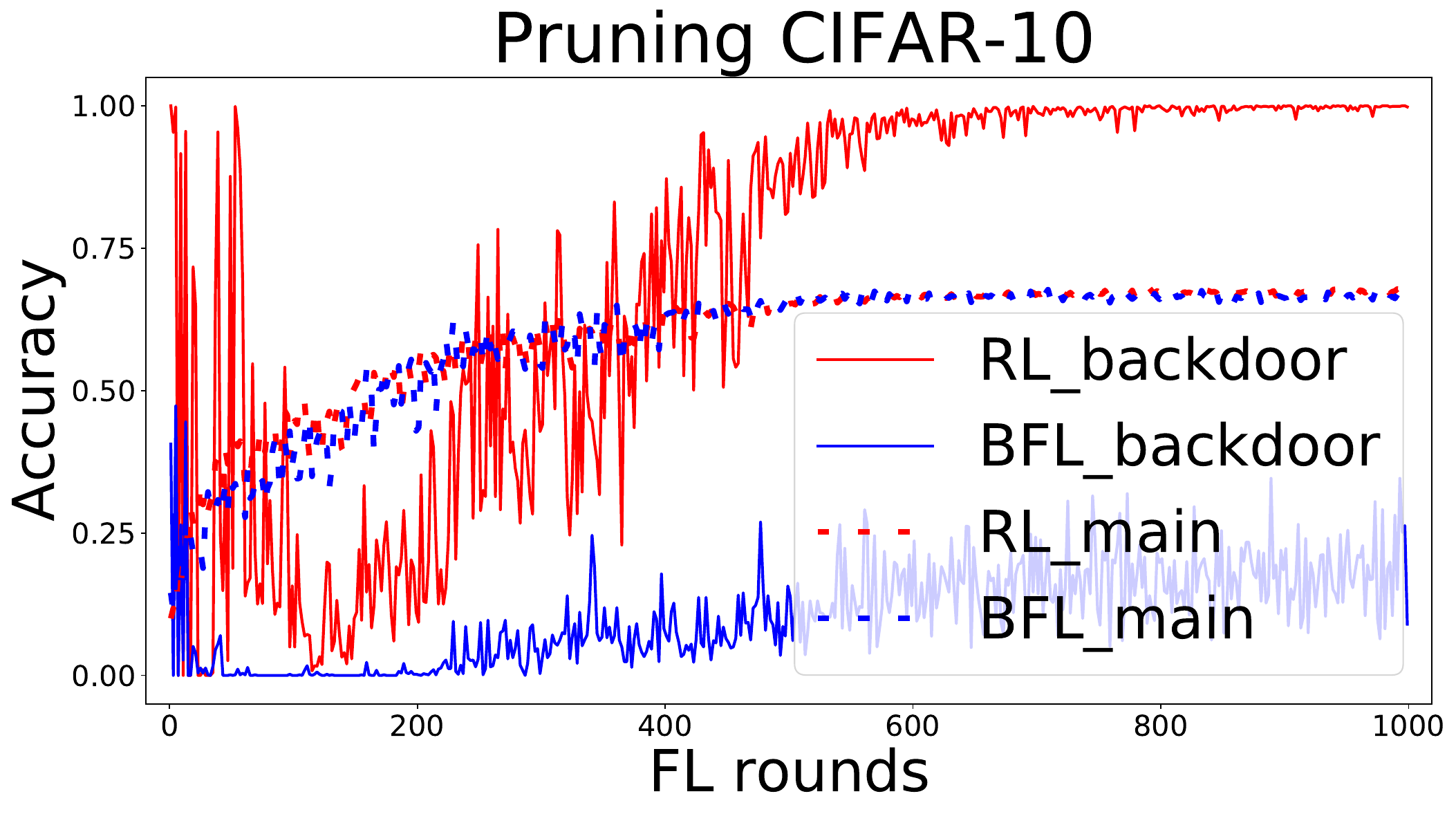}
          }
          }}
         &
    \hspace{-1em}
    \adjustbox{valign=m}{{%
    \subfigure[]{
          \centering
          \includegraphics[width=.31\textwidth]{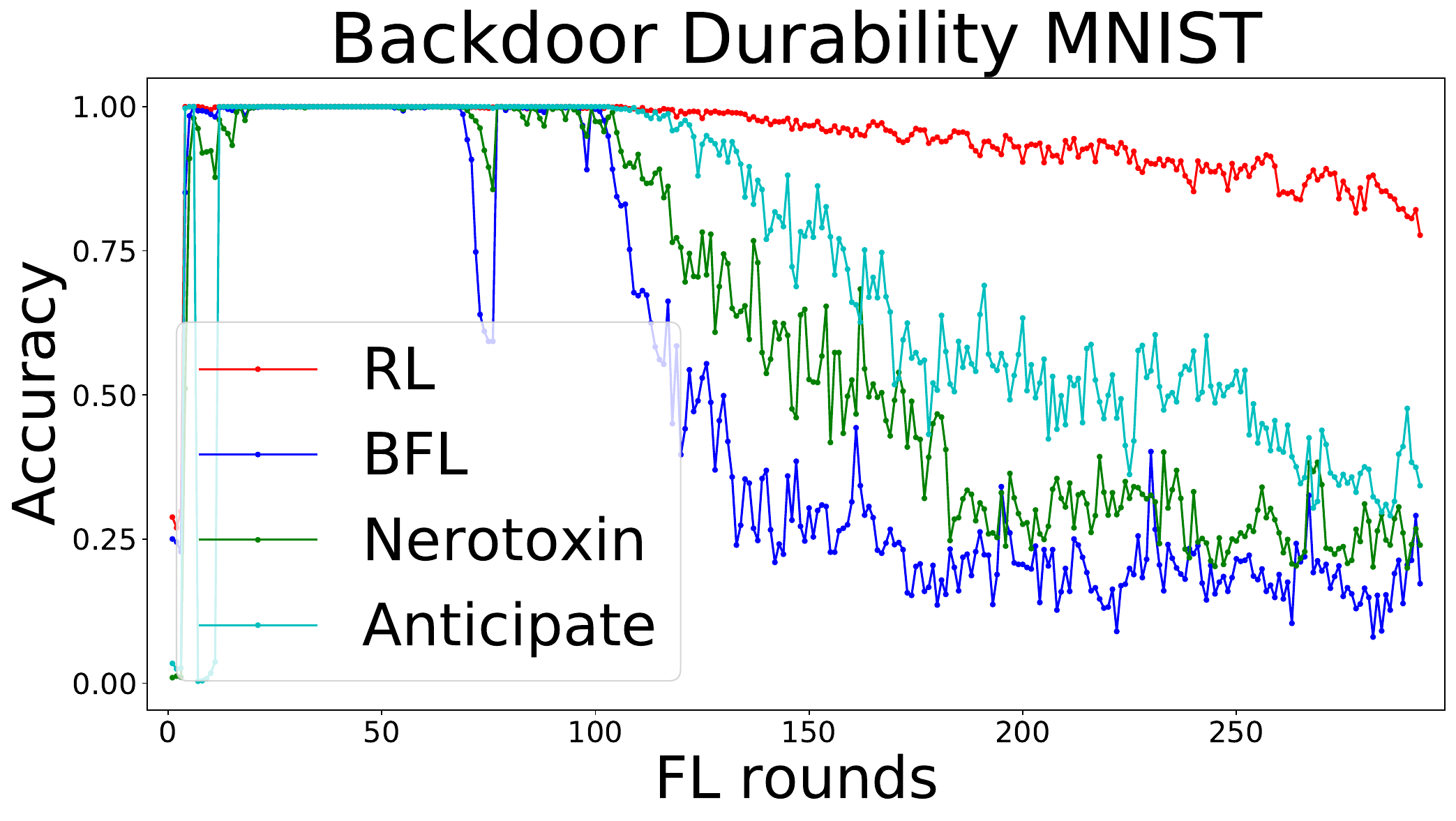}
          }
          }}
          \\
    \end{tabular}
    \caption{\small{ (a)-(b) Global model accuracy for the backdoor and main tasks during FL training, under BFL and RL attacks, and Neuron Clipping and Pruning defenses. (c) 
    Backdoor durability of the RL attack, BFL, Neurotoxin, and Anticipate. Attacks are applied during the first 100 FL rounds only.
    }}\label{fig:results}
\end{figure*}

\paragraph{Post-Training Defenses.} Figures~\ref{fig:results}(a) and~\ref{fig:results}(b) show the test accuracy of the global model with respect to the backdoor and main tasks under BFL and our RL attack when the server uses FedAvg as the training stage aggregation rule and Neuron Clipping or Pruning as the post-training defense. 
We envision that a post-training defense can be applied to an intermediate global model during FL training, although its performance can be unstable before the main task is close to convergence.  
Thus, we include it in computing the reward function during RL training and in evaluating the test accuracy of an intermediate global model. 
We observe that both defense methods can effectively decrease the backdoor accuracy for BFL. However, our RL attack bypasses both of them by incorporating their influence into its reward function. 

\paragraph{Durability.} In Figure~\ref{fig:results}(c), we compare the backdoor durability of our RL attack, BFL, Neurotoxin, and Anticipate, with FedAvg as the aggregation rule and no post-training defense applied. Attacks are applied during the first 100 FL rounds only. We observe that all attack methods reach high backdoor accuracy ($\sim$$100\%$) while the attack happens. After the first 100 FL, our RL attack shows the slowest accuracy decay among all attacks. At round 300, our RL attack still maintains $80\%$ backdoor accuracy, while other attacks' backdoor accuracy is all below $40\%$. The reason behind the high durability of our attack is that it optimizes for a long-term attack objective by simulating the future behavior of the FL system under attack. 

\section*{Acknowledgments}
This work was supported in part by NSF award CNS-2146548. We thank the three
anonymous reviewers for their constructive feedback.

\bibliography{iclr2023_conference}
\bibliographystyle{iclr2023_conference}

\appendix
\section*{Appendix}
\section{The Backdoor Attack Algorithm}

\begin{algorithm}[!ht]
\begin{algorithmic}
\STATE \textbf{Server Input: }Initial weight $w^0_g$, set of all clients $[K]$, all clients' local datasets $\{D_1,D_2,\cdots, D_K\}$, subsampling rate $\kappa$, aggregation rule $Aggr$, post-training defense $h(\cdot)$
\STATE \textbf{Client Input: }Local minibatch size $B$, local step size $\eta$, local iteration number $E$, number of global training steps $T$, local search policy $\pi_1$, model crafting policy $\pi_2$\
\STATE \textbf{Server Output: }$\widehat{w}^{T}_g$
\begin{multicols}{2}
\STATE \textbf{Server executes:}
\begin{ALC@g}
\FOR{$t=0$ to $T-1$}
\STATE $\mathcal{S}^t \leftarrow$ randomly sample $\kappa \cdot K$ clients 
\STATE Broadcast $w_g^t$ to all clients in $\mathcal{S}^t$
\STATE Wait for all model updates $\{g^{t}_{k}\}_{k\in \mathcal{S}^t}$ 
\STATE Update $w^{t+1}_g \leftarrow w^{t}_g - Aggr(\{g_k^t\}_{k \in \mathcal{S}^t})$
\ENDFOR
\STATE $\widehat{w}^{T}_g \leftarrow h(w^{T}_g)$
\end{ALC@g}

\vspace{2ex}
\STATE {\bf BenignUpdate}$(k, w)$:
\begin{ALC@g}
\STATE $w_0 \leftarrow w$
\FOR{local epoch $e=1$ to $E$ }
\STATE Sample a minibatch $b$ of size $B$ from $D_k$
\STATE $w \leftarrow w- \eta \frac{1}{B} \sum_{(x,y) \in b}
\nabla_w\ell(w, (x,y))$
\ENDFOR
\STATE $g \leftarrow w_0-w$
\STATE Send $g$ to server 
\end{ALC@g}

\vspace{2ex}
\STATE {\bf DoubleWhammyUpdate}$(k, w)$:
\begin{ALC@g}
\STATE Malicious devices communicate to get $|\mathcal{A}|$
\STATE $\rho, B', E', \eta' \leftarrow \pi_1(w, |\mathcal{A}|)$
\STATE $\alpha, \beta \leftarrow \pi_2(w, |\mathcal{A}|)$
\STATE $D'_k \leftarrow$ randomly poison local dataset $D_k$ with poison ratio $\rho$
\STATE $w_0 \leftarrow w$ and $w' \leftarrow w$
\FOR{local epoch $e=1$ to $E'$ }
\STATE Sample a minibatch $b'$ of size $B'$ from $D'_k$
\vspace{-2.5ex}
\STATE $w' \leftarrow w' - \eta'\frac{1}{B'} \sum_{(x,y) \in b'}
\nabla_w\ell(w, (x,y))$
\ENDFOR
\STATE $\tilde{g} \leftarrow w_0-w'$
\FOR{local epoch $e=1$ to $E$ }
\STATE Sample a minibatch $b$ of size $B$ from $D_k$
\STATE $w \leftarrow w- \eta \frac{1}{B} \sum_{(x,y) \in b}
\nabla_w\ell(w, (x,y))$
\ENDFOR
\STATE $g \leftarrow w_0-w$
\FOR{each layer $l$ of $\tilde{g}$ {\bf in parallel}}
\STATE Select top-$\alpha\%$ coordinates of $|\tilde{g}[l]-g[l]|$
\ENDFOR
\FOR{each coordinate $i$ of all selected coordinates {\bf in parallel}}
\STATE $\tilde{g}[i] \leftarrow \tilde{g}[i] - \beta (\tilde{g}[i]-g[i])$
\ENDFOR
\STATE Send $\tilde{g}$ to server 
\end{ALC@g}
\end{multicols}
\end{algorithmic}
 \caption{Federated Learning with Double Whammy Backdoor Attack}
 \label{alg:fl}
\end{algorithm}

\section{Experiment Setup}
\label{sec:setup}
\begin{figure*}
    \centering
    \includegraphics[width=.99\textwidth]{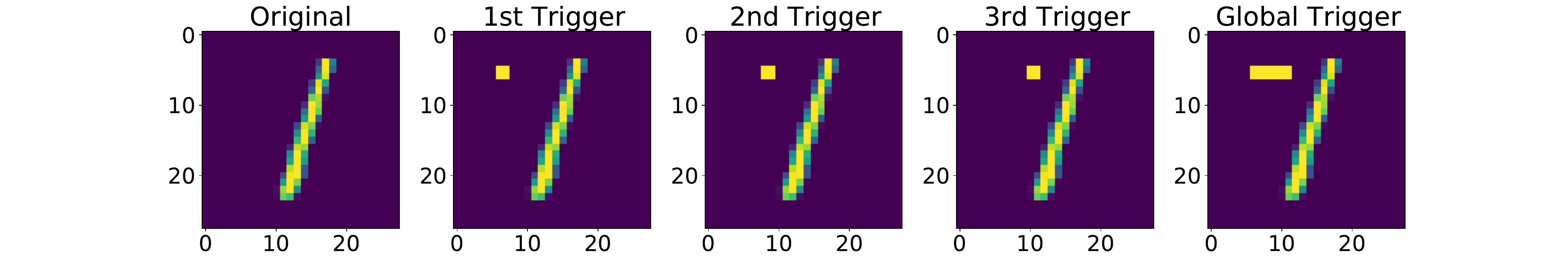}
     \caption{\small{MNIST 
     backdoor trigger patterns. 
     The global trigger is considered the default poison pattern and is used for backdoor accuracy evaluation. The sub-triggers are used by DBA only.}}
    \label{fig:mnist_dba}
\end{figure*}

\begin{figure*}
    \centering
    \includegraphics[width=.99\textwidth]{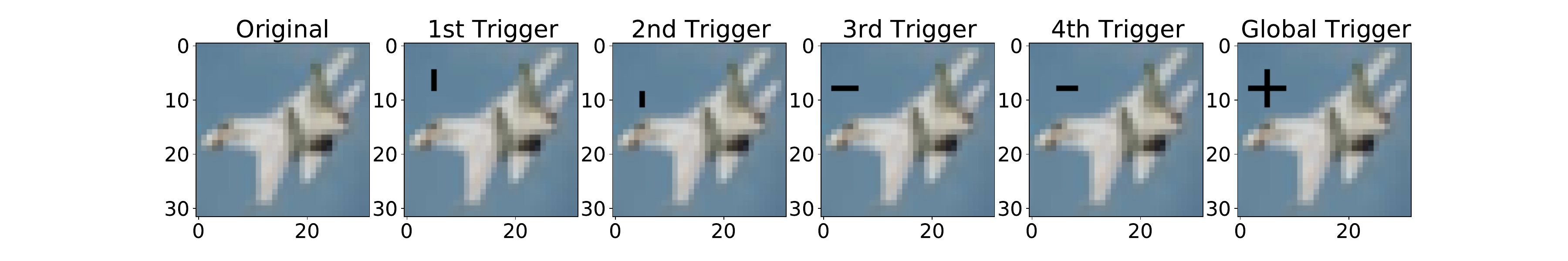}
    \caption{\small{CIFAR-10 backdoor trigger patterns.
    The global trigger is considered the default poison pattern and is used for backdoor accuracy evaluation. The sub-triggers are used by DBA only.}}
    \label{fig:cifar10_dba}
\end{figure*}

\paragraph{Datasets.}We consider two datasets: MNIST \citep{lecun1998gradient} and CIFAR-10 \citep{Krizhevsky2009LearningML}, and $i.i.d.$ local data distributions, where we randomly split each dataset into $K$ groups, each with the same number of training samples. MNIST includes 60,000 training examples and 10, 000 testing examples, where each example is a 28$\times$28 grayscale image, associated with a label from 10 classes. CIFAR-10 consists of 60,000 color images in 10 classes of which there are 50, 000 training examples and 10,000 testing examples. We consider the trigger patterns shown in Figure~\ref{fig:mnist_dba} and Figure~\ref{fig:cifar10_dba}, where the goal is to misclassify digit 1 to digit 7 for MNIST and airplane class to truck class for CIFAR-10. The default poison ratio is 0.5 in both cases. We consider both a global trigger and a set of sub-triggers for each dataset as shown in the figures, similar to DBA~\cite{xie2019dba}. The global trigger is used as the default poison trigger pattern and the pattern for evaluating the test accuracy of all backdoor attacks, while the sub-triggers are only used by DBA.
 
\paragraph{Federated Learning Setting.}We use the following default parameters for the FL environment: local minibatch size = 128, local iteration number = 1, learning rate = 0.05, number of workers = 100, number of attackers = 5, subsampling rate = $10\%$, and the number of FL training rounds = 500 (resp. 1000) for MNIST (resp. CIFAR-10). For MNIST, we train a neural network classifier consisting of two 5$\times$5 convolutional filter layers with ReLU activations followed by two fully connected layers and softmax output. For CIFAR-10, we use the ResNet-18 model~\citep{he2016deep}.
We implement the FL model with PyTorch~\citep{paszke2019pytorch} and run all the experiments on the same 2.30GHz Linux machine with 16GB NVIDIA Tesla P100 GPU. 
We use the cross-entropy loss as the default loss function and stochastic gradient descent (SGD) as the default optimizer. For all the experiments, we fix the random seeds of subsampling for fair comparisons.

\paragraph{Baselines.}We compare our RL-based attack (RL) with the state-of-the-art backdoor FL attack methods: BFL~\citep{bagdasaryan2020backdoor} without model replacement, DBA~\citep{xie2019dba} where each  selected attacker randomly chooses a sub-trigger as shown in Figures~\ref{fig:mnist_dba} and~\ref{fig:cifar10_dba}, PGD attack~\citep{wang2020attack} with a projection norm of 0.05, 
Neurotoxin~\citep{zhang2022neurotoxin} with top-100 ($\sim$$1\%$) parameters for masking at each layer. 
and Anticipate~\citep{wen2022thinking} with 5 look-ahead steps. 
For training stage defenses, we consider four aggregation-based methods: FedAvg~\citep{mcmahan2017communication}, Krum~\citep{blanchard2017machine}, Median~\citep{yin2018byzantine}, and norm-bounding~\citep{sun2019can} (with a norm bound of 0.02). For post-training defenses, we consider the backdoor mitigation methods in~\cite{wang2022universal} (Neuron Clipping) and~\cite{wu2020mitigating} (Pruning). We use the  original clipping thresholds in~\cite{wang2022universal} and set the default pruning number to 256.    

\paragraph{Reinforcement Learning Setting.}In our RL-based attack, since both the action space and state space are continuous, we choose the state-of-the-art Twin Delayed DDPG (TD3)~\citep{fujimoto2018addressing} algorithm to alternatively train the local search policy for 10,000 steps and the model crafting policy for 10,000 steps in each iteration, with 80,000 total training steps (i.e., the number of training iterations is 4). We implement our simulated environment with OpenAI Gym~\citep{1606.01540} and adopt OpenAI Stable Baseline3~\citep{raffin2021stable} to implement TD3.
The RL training parameters are described as follows: 
the number of FL rounds = 500 rounds, policy learning rate = 0.001, the policy model is MultiInput Policy, batch size = 256, and $\gamma$ = 0.99 for updating the target networks.
The default $\lambda=0.5$ when calculating the rewards. For Median (resp. Krum), we set $\lambda=0.4$ (resp. $\lambda=0.375$).

As described in Section~\ref{sec:DWBA}, we split the attackers' action into two sub-actions $a^t_1$ and $a^t_2$ corresponding to the local search step and the model crafting step, respectively. Let $A_1$ and $A_2$ denote the action spaces in the two steps and define the local search policy and the model crafting policy as $\pi_1: S \rightarrow A_1$ and $\pi_2: S \rightarrow A_2$, respectively.
Instead of training $\pi=(\pi_1,\pi_2)$ simultaneously, we implement an alternative training structure in the simulated environment, where we alternatively train $\pi_1$ and $\pi_2$ while fixing the other policy. Figure~\ref{fig:more_results}(b) shows the advantage of this approach compared with simultaneous training.


\section{More Experiment Results}

\subsection{Attack Performance}

\begin{figure*}[t]
    \centering
    \begin{tabular}{ccc}
    \adjustbox{valign=m}{{%
    \subfigure[]{
          \centering
          \includegraphics[width=.46\textwidth]{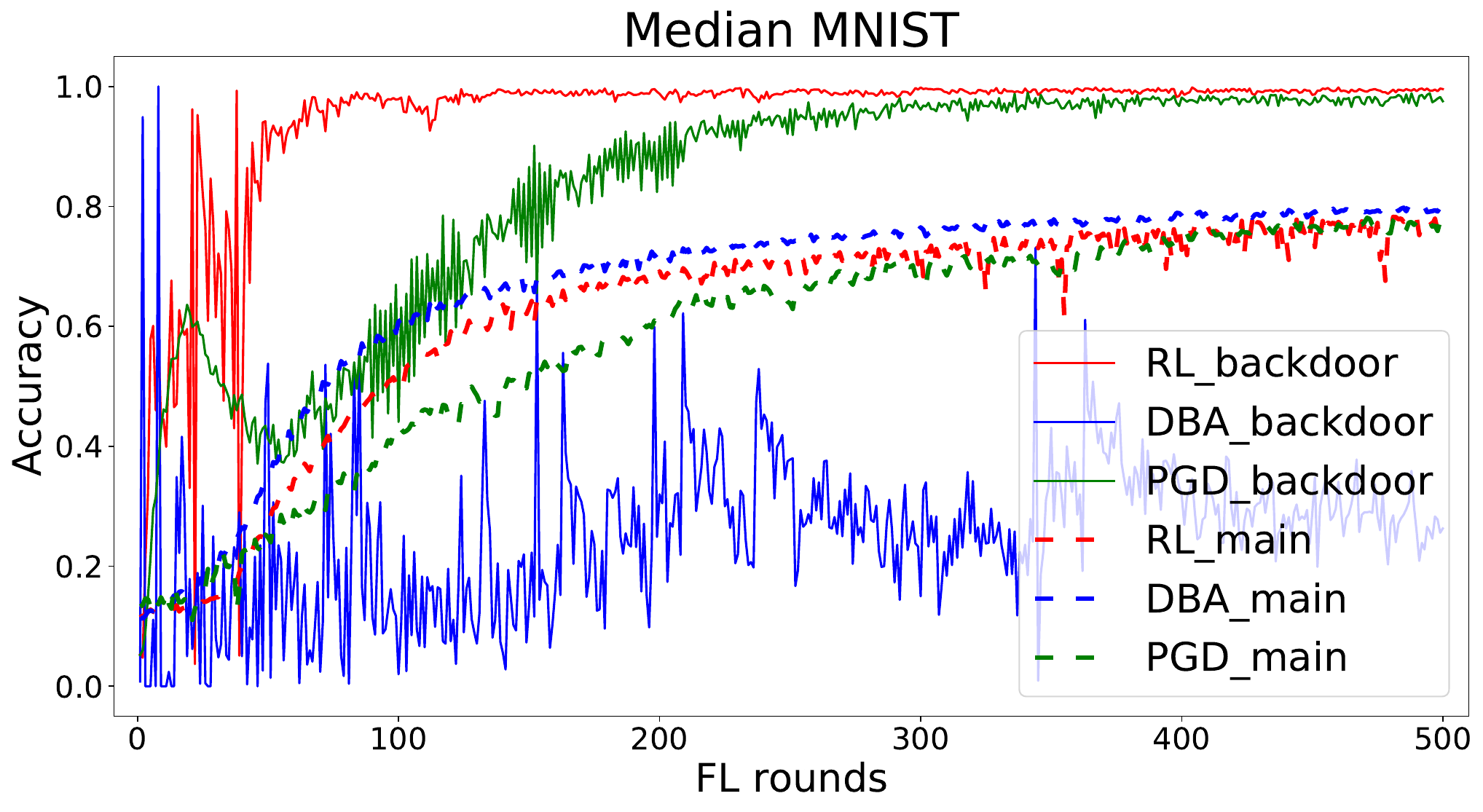}
          }
          }}
         &
    \adjustbox{valign=m}{{%
    \subfigure[]{
          \centering
          \includegraphics[width=.46\textwidth]{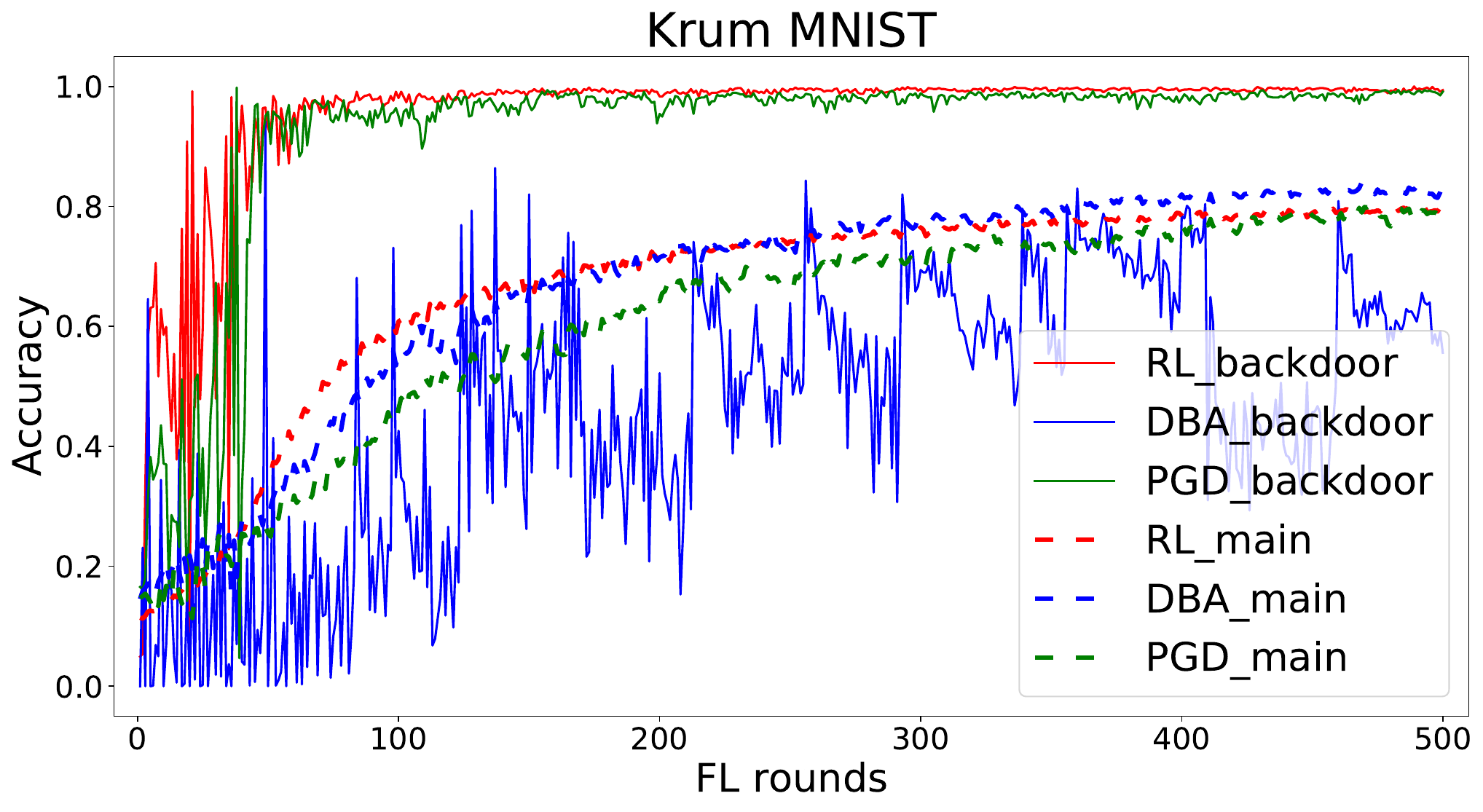}
          }
          }}
         &
          \\
    \end{tabular}
    \caption{\small{A comparison of the global model accuracy for the backdoor task  and the main task during FL training, under RL attack, DBA, and PGD attack with Krum and Median defenses. All parameters are set as default. 
    }}\label{fig:train-stage3}
\end{figure*}

\begin{figure*}[t]
    \centering
   \begin{tabular}{ccc}
    \adjustbox{valign=m}{{%
    \subfigure[]{
          \centering
          \includegraphics[width=.46\textwidth]{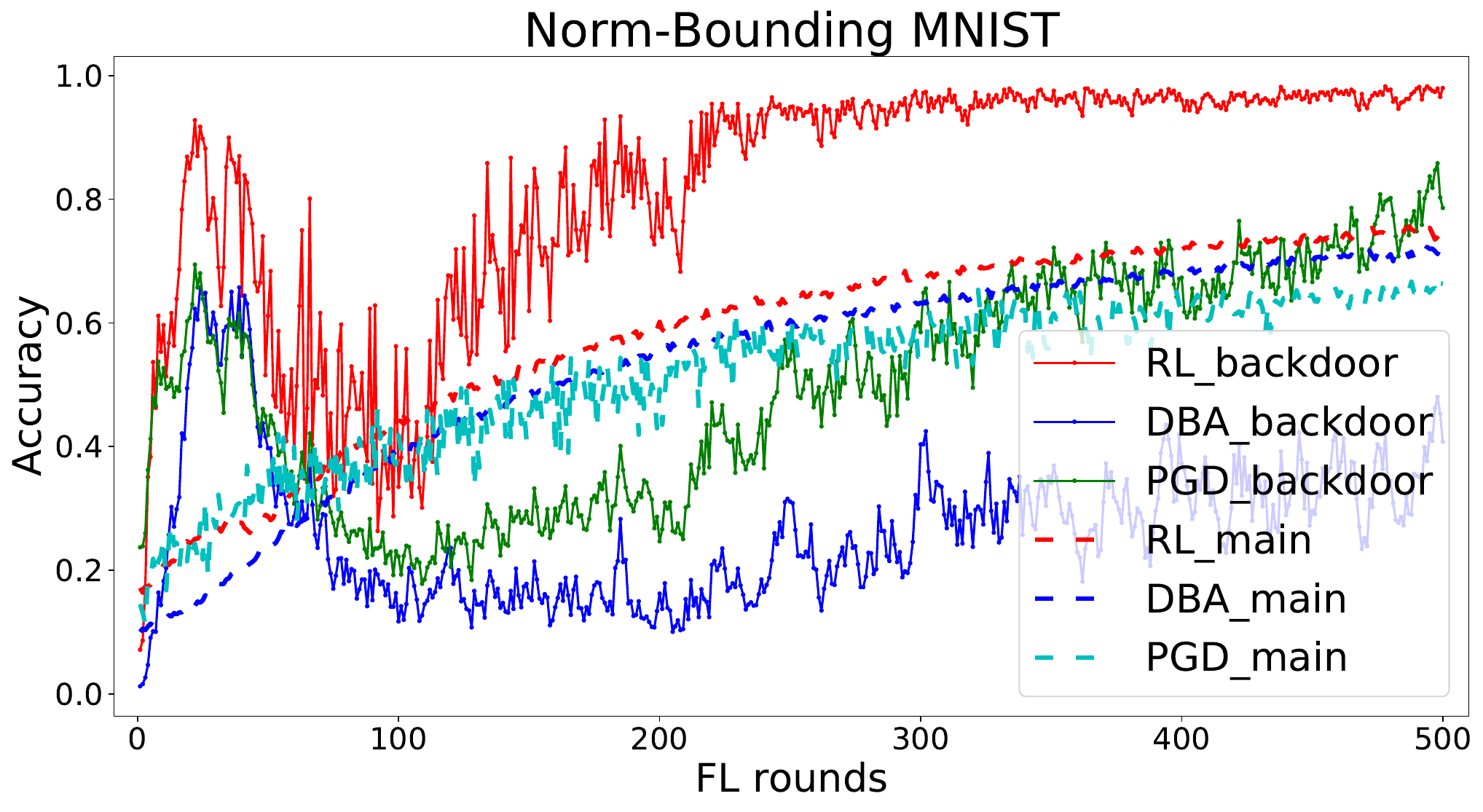}
          }
          }}
         &
    \adjustbox{valign=m}{{%
    \subfigure[]{
          \centering
          \includegraphics[width=.46\textwidth]{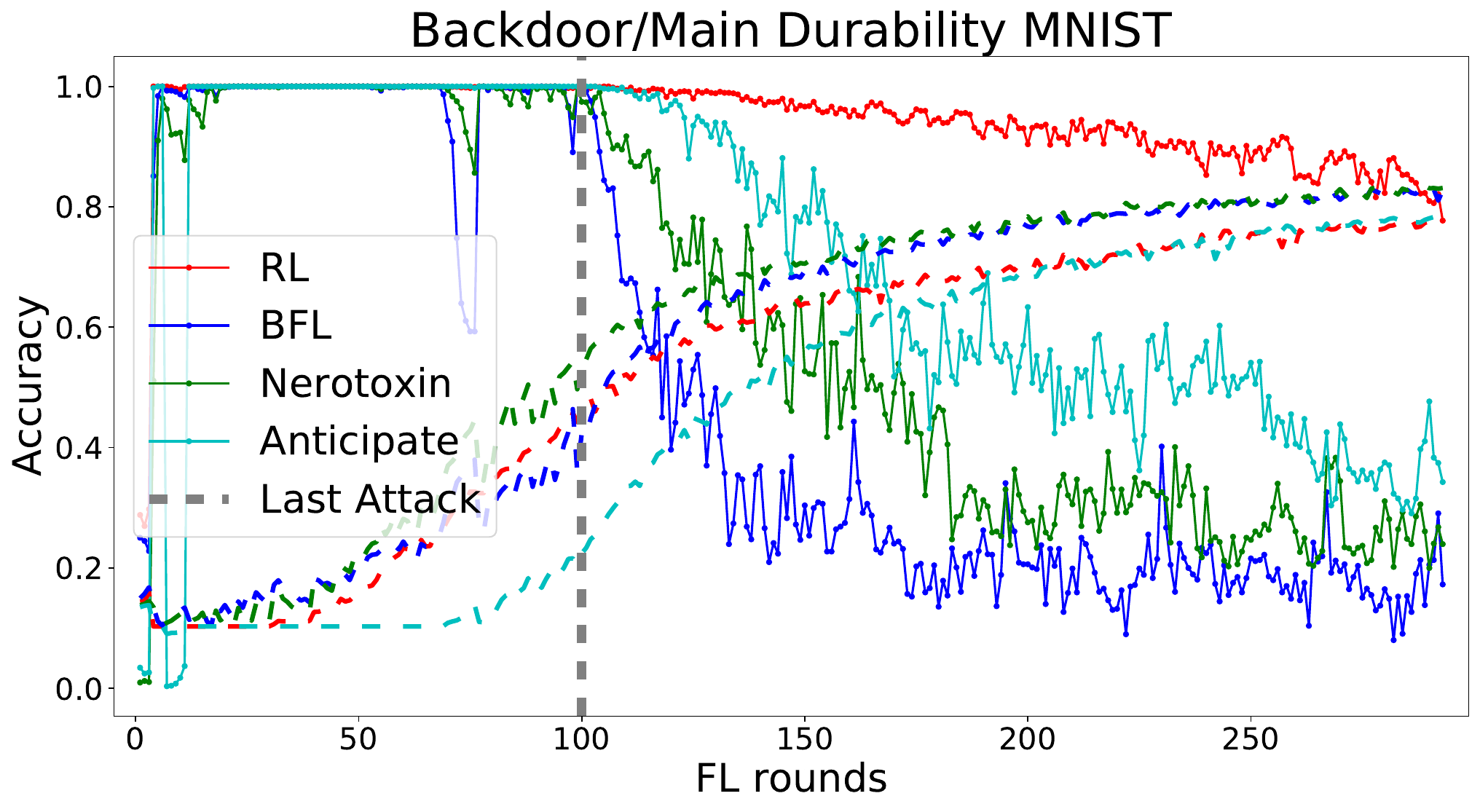}
          }
          }}
         &
          \\
    \end{tabular}
    \caption{\small{(a) A comparison of the global model accuracy for the backdoor task and the main task during FL training, under RL attack, BFL, PGD attack, and norm-clipping defense. (b) A comparison of backdoor attack durability (solid line) and the corresponding main task accuracy (dashed line) of the RL attack, BFL, Neurotoxin, and Anticipate. Attacks are applied during the first 100 FL rounds only.
    }}\label{fig:train-stage2}
\end{figure*}

\paragraph{More Baseline Attacks.} In Figure~\ref{fig:train-stage3}, we compare the backdoor task and main task accuracy during FL training under our RL-based attack, DBA, and PGD attack, when Krum or Median is considered as the defense mechanism. Among the three attacks, the RL attack reaches $\sim$$100\%$ backdoor accuracy within 100 FL rounds. The backdoor accuracy of DBA is unstable due to the introduction of subsampling, which is ignored in the original paper. Further, its attack performance stays at a low level during the whole FL training in our setting. This is mainly because each DBA attacker uses a randomly sampled sub-trigger (see Figure~\ref{fig:mnist_dba}) while baseline attackers all use the same global trigger. The latter is more powerful given the same number of attackers.
Among the three baselines, PGD achieves relatively better performance compared with BFL (see Figure~\ref{fig:train-stage}) and DBA. However, our RL-based attack still outperforms the PGD attack as it utilizes adaptive actions based on a long-term objective. The attack performance of all attacks typically grows faster under Krum compared with Median due to the fact that under Krum, an attacker's input will completely replace the actual global model once it is chosen by the server. 
Figure~\ref{fig:train-stage2}(a) gives similar results under the norm-bounding defense. Compared with DBA, PGD attack achieves better backdoor accuracy but results in relatively lower main task accuracy, due to the fixed projection bound and other hyperparameters chosen. Our RL attack outperforms all baselines for both backdoor task and main task accuracy.

\begin{figure*}[!t]
    \centering
   \begin{tabular}{ccc}
    \adjustbox{valign=m}{{%
    \subfigure[]{
          \centering
          \includegraphics[width=.31\textwidth]{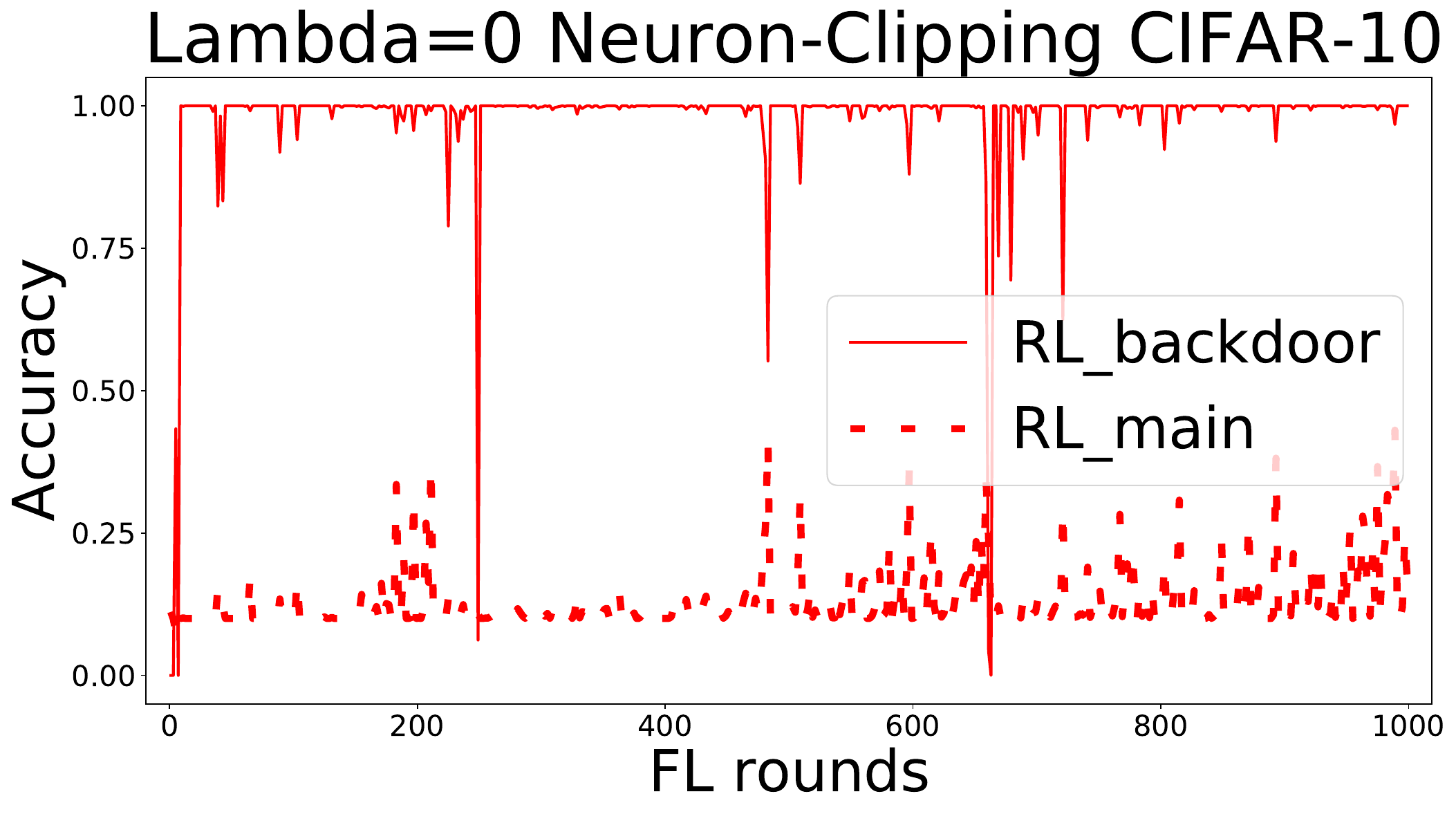}
          }
          }}
         &
    \hspace{-1em}
    \adjustbox{valign=m}{{%
    \subfigure[]{
          \centering
          \includegraphics[width=.31\textwidth]{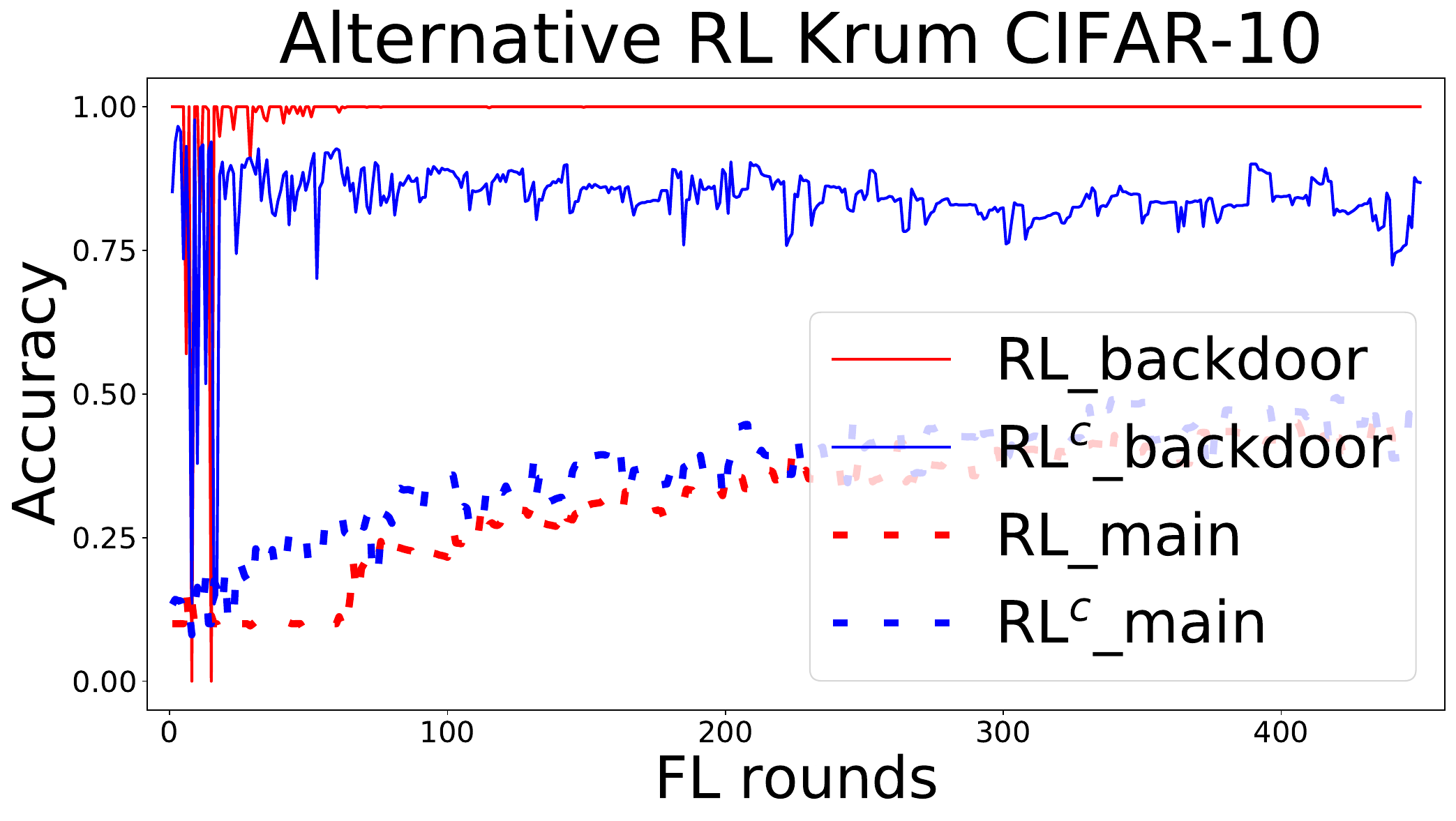}
          }
          }}
         &
    \hspace{-1em}
    \adjustbox{valign=m}{{%
    \subfigure[]{
          \centering
          \includegraphics[width=.3\textwidth]{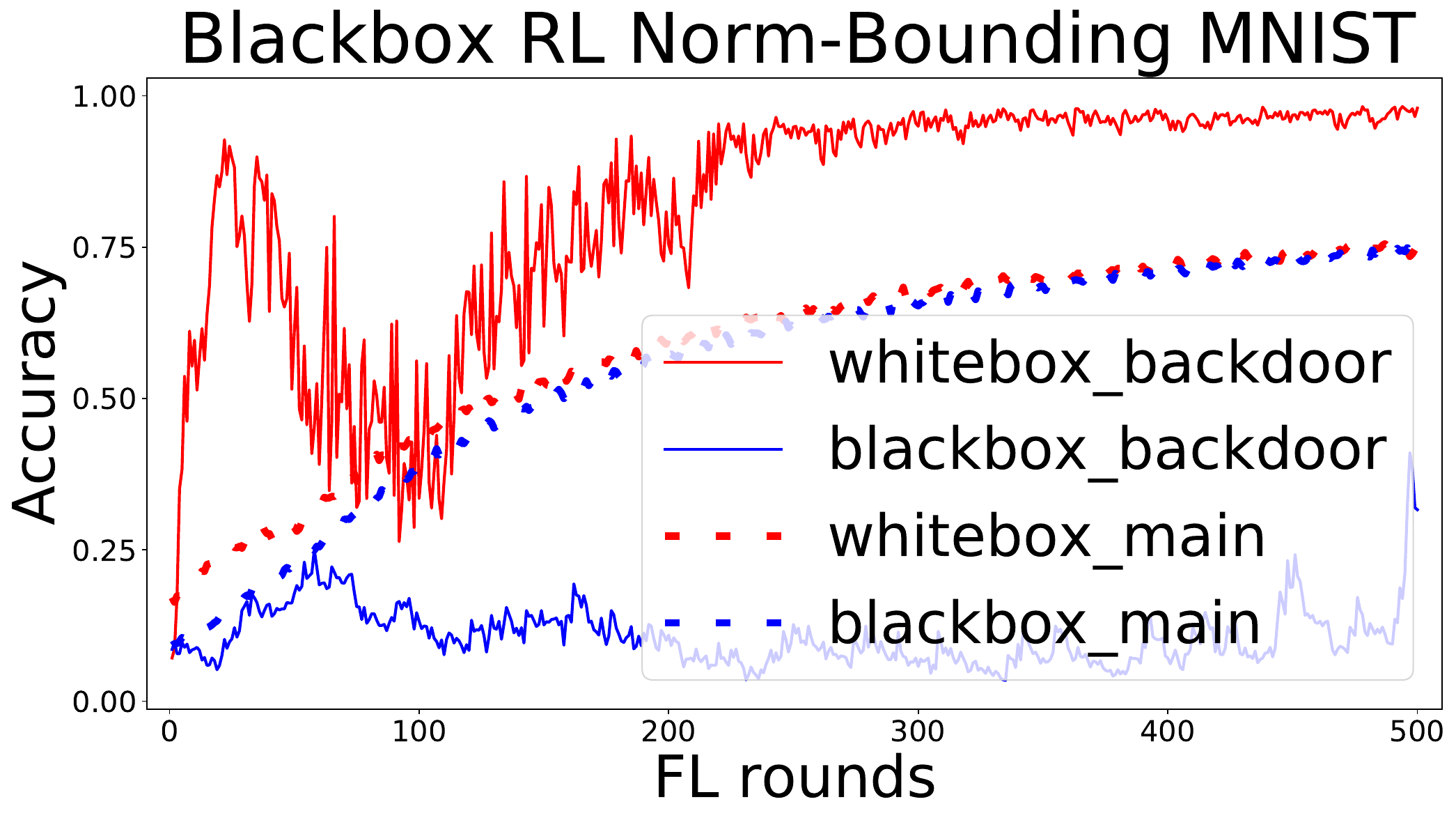}
          }
          }}
          \\
    \end{tabular}
    \caption{\small{A comparison of the global model accuracy for the backdoor task and the main task during FL training, under (a) RL attack, BFL, and post-training Neuron Clipping defense; (b) RL attacks based on alternative training (RL) and simultaneous training (RL$^c$), where the lengths of training are both 80,000 steps, and Krum defense; (c) white-box vs. black-box RL attack under the norm-bounding defense.
    }}\label{fig:more_results}
\end{figure*}

\paragraph{Additional Durability Result.} Figure~\ref{fig:train-stage2}(b) shows the accuracy of both backdoor and main tasks when the attacker only attacks during the first 100 FL rounds. We observe that Anticipate achieves better backdoor durability than BFL and Neorontoxin but at the cost of a slower growth rate of the main task accuracy. Our RL attack obtains the highest backdoor durability while maintaining good main task accuracy. The latter can be further improved by tuning $\lambda$ and the discount factor $\gamma$ during training. 

\subsection{Ablation Studies}

\paragraph{Impact of $\lambda$.} 

Figure~\ref{fig:more_results}(a) shows the backdoor and main task accuracy when $\lambda=0$. Compared with Figure~\ref{fig:results}(a) where $\lambda=0.5$, we observe that the backdoor accuracy for the RL attack further improves as expected while the main task accuracy becomes significantly worse as it is completely ignored in the reward function. On the other hand, the performance of BFL is independent of $\lambda$ (as it only uses the poison ratio to obtain such a tradeoff). This indicates the importance of choosing a proper $\lambda$ (or poison ratio) to maximize the backdoor attack performance while maintaining good main task accuracy. As it is time consuming to search for such a $\lambda$ manually, a promising direction is to consider constrained RL with the desired main task accuracy as a constraint, which is left to our future work. 

\paragraph{Efficiency of Alternative Training.} 
Figure~\ref{fig:more_results}(b) compares a policy trained using the alternative training approach (called RL) and a policy trained using the simultaneous training approach (called RL$^c$), 
where both policies are trained for 80,000 time steps in total. For alternative training, we train the local search policy for 10,000 steps and the model crafting policy for 10,000 steps in each iteration, with 80,000 total training steps (i.e., the number of training iterations is 4). 
We observe that the former 
achieves nearly $100\%$ backdoor accuracy while the latter only reaches a backdoor accuracy around $80\%$, indicating that the simultaneous training does not converge over $80,000$ steps.

\paragraph{Importance of Knowing FL System Parameters.}
In this work, we focus on the commonly considered white-box attack setting and assume that the RL attacker has prior knowledge about the FL environment, including the server's defense mechanism, the local training method, the number of devices, and the subsampling rate. Figure~\ref{fig:more_results}(c) gives the result of a naive black-box RL attack, where the system adopts the norm-bounding defense (with the default clipping bound), while the attacker does not know that and instead assumes that FedAvg is used as the aggregation rule and there is no post-training defense. As shown in the figure, the black-box attack fails to compromise the FL system. This is because with the wrong ``world model'', the attacker tends to learn an aggressive attack policy, resulting in malicious model updates that are significantly different from benign updates and can be easily filtered out by the defense. Intuitively, the attacker should be more conservative in the black-box setting by considering a worst-case defense scenario so that it could achieve certain level of attack performance even under weaker defenses. Further, the attacker should constantly adapt to the unknown environment by updating its policy using the real-time feedback obtained during online FL training. How to design an effective RL-based backdoor attack in the black-box setting is an interesting open problem and is left to our future work. 
 

\end{document}